# Un modèle de génération de dilemmes de prohibition et d'obligation en environnement virtuel

▶ Azzeddine BENABBOU, Domitile LOURDEAUX, Dominique LENNE (Sorbonne université, Université de technologie de Compiègne, CNRS UMR 7253 Heudiasyc)

■ RÉSUMÉ • Dans le cadre du projet Maccoy Critical, nous souhaitons former les apprenants à la gestion des situations critiques telles que les dilemmes. Un dilemme correspond à une situation où il n'existe pas de bonne solution, c'est-à-dire à une situation qui mène à des conséquences négatives dans tous les cas. Notre objectif est d'utiliser des modèles de connaissances pour en extraire les propriétés nécessaires à l'émergence de dilemmes. Notre approche consiste pour cela à développer un système de scénarisation capable de générer des dilemmes sans avoir à les écrire à l'avance. Dans cet article nous présentons cette approche et exposons une preuve de concept appliquée à la conduite automobile.

■ *ABSTRACT • Under the project Maccoy Critical, we would like to train individuals, in virtual environments, to handle critical situations such as dilemmas. These latter refer to situations where there is no "good" solution. In other words, situations that lead to negative consequences whichever choice is made. Our objective is to use Knowledge Models to extract necessary properties for dilemmas to emerge. To do so, our approach consists in developing a Scenario Orchestration System that generates dilemma situations dynamically without having to write them beforehand. In this paper we present this approach and expose a proof of concept of the generation process.*

# 1. Introduction

Il arrive lors de certains évènements de la vie professionnelle que des agents soient confrontés à des situations critiques. Celles-ci sont définies comme étant des situations dynamiques et complexes où des facteurs tant internes (p. ex. état de fatigue, manque de concentration) qu'externes (conditions climatiques dégradées, incompétences des collègues) à l'individu font qu'elles s'écartent des situations a priori maitrisées. Pour gérer ce genre de situations, et éviter par la suite des conséquences parfois désastreuses, il est nécessaire d'entrainer les agents concernés à réagir dans ce type de situations. Ils doivent être confrontés pour cela à une variété de scénarios où ils devront comprendre leur environnement et agir, parfois en urgence, développant ainsi les compétences nécessaires.

## 1.1. Les environnements virtuels pour la formation

La production de situations d'entrainement n'est pas toujours évidente. En effet, la dangerosité, le manque de ressources et les problèmes d'accessibilité font que les scénarios ne sont pas toujours reproductibles en conditions réelles. La réalité virtuelle peut pallier ce problème en offrant des outils et des techniques de simulation permettant d'immerger les personnes dans des environnements virtuels fidèles aux situations réelles. Elle permet un apprentissage situé et constructiviste : l'apprenant restructure ses connaissances par l'expérience, par la confrontation à des situations variées, nouvelles et plus ou moins courantes. Elle permet de le confronter à des situations d'un niveau de criticité variable en fonction de ses actions et de son profil. Ces situations doivent être suffisamment difficiles tout en restant abordables. Elles doivent permettre de confronter les compétences acquises, mais aussi de mettre en jeu des compétences nouvelles et proches de celles acquises pour favoriser le développement (Vygotsky, 1978). Pour supporter ce type d'apprentissage, il est nécessaire de générer un large spectre de scénarios. L'écriture de tels scénarios est un travail important qui conduit, lors du passage à l'échelle, à ce qu'on appelle l'*Authoring Bottleneck* (Spierling et Szilas, 2009). Il est alors nécessaire de mettre en place des systèmes de scénarisation permettant de créer des environnements adaptables, sans avoir à définir explicitement l'intégralité des scénarios possibles. Pour conserver la liberté d'action de l'utilisateur et assurer l'adaptabilité des scénarios, nous étudions l'hypothèse qu'il est possible de générer dynamiquement et automatiquement des situations critiques à partir de modèles de connaissances qui sous-tendent la simulation. Nous nous intéressons au processus de scénarisation de situations critiques en environnement virtuel. La scénarisation est un processus comprenant à la fois la spécification du ou des déroulements possibles ou souhaitables de la simulation, et le contrôle (exécution et/ou suivi et correction) du déroulement des événements en temps interactif. Un système de scénarisation est composé d'un ou plusieurs langages de scénarisation permettant de modéliser le contenu scénaristique et/ou les objectifs scénaristiques, et d'un moteur de scénarisation permettant de gérer de manière dynamique la réalisation du scénario (Barot, 2014).

### 1.2. Contexte et objectifs

Ces travaux de recherche s'intègrent dans le cadre du projet national Maccoy Critical (Burkhardt *et al.*, 2016) dont l'objectif est l'étude et l'amélioration des dispositifs de simulation et de réalité virtuelle pour la formation. En particulier, nous nous intéressons à la génération de situations critiques. Le but est de confronter les apprenants à ces situations afin de leur faire acquérir des compétences non-techniques. Parmi ces situations, on trouve les dilemmes. Un dilemme est une situation où l'individu est dans « *l'obligation de choisir entre deux partis qui comportent l'un et l'autre des inconvénients* »[1]. Dans la vie réelle, ces situations sont rencontrées quotidiennement dans différents domaines. Nous nous intéressons à la médecine et à la conduite automobile. En médecine, les infirmières sont confrontées quotidiennement à des dilemmes éthiques. Dans son article (Lecomte, 2006), Lecomte explique pourquoi les infirmières devraient avoir une réflexion éthique. Elle cite en exemple le fait de pouvoir garder une vue holistique de la personne, d'être capable de négociation et de compromis, mais aussi de prendre de la distance et d'avoir un autre regard pour prendre des décisions. Les dilemmes sont aussi présents dans d'autres domaines.

En conduite automobile, le cas de figure le plus célèbre est celui où le conducteur doit choisir entre percuter des piétons ou se sacrifier lui-même ainsi que ses passagers. Ce dilemme fait l'objet de plusieurs études comme celle de (Bonnefon *et al.*, 2016) dont l'objectif est d'étudier et de déterminer quel comportement un véhicule autonome devrait adopter dans ce cas-là.

[1] Version en ligne du dictionnaire Larousse (année 2018).

La conception d'environnements virtuels pour l'entrainement ne concerne pas que les informaticiens auteurs des logiciels de simulation. Elle implique aussi d'autres personnes à savoir des ergonomes, des formateurs, des experts du domaine, etc. Généralement, ces personnes manipulent des modèles de connaissances différents et spécifiques à leur domaine d'expertise (p. ex. modèle de tâches, modèle du monde). Notre objectif est de concevoir un système de scénarisation capable de raisonner sur ces différents modèles afin de faire émerger des dilemmes.

## 2. Etat de l'art

### 2.1.1. Les dilemmes en simulation

Plusieurs travaux de l'Institute for Creative Technologies (ICT) (Rickel *et al.*, 2002) implémentent le scénario suivant : jouant le rôle d'un lieutenant de l'armée américaine, l'utilisateur doit venir en renfort à un peloton en difficulté. Pendant le trajet, le lieutenant et sa troupe passent par un village où ils croisent un garçon blessé en situation critique. Un véhicule de la troupe du lieutenant est impliqué dans l'accident. Par conséquent, le lieutenant est confronté à l'alternative suivante : poursuivre son chemin pour venir en renfort à son peloton ou sécuriser une zone d'atterrissage pour permettre à un hélicoptère médical de se poser et venir en secours au garçon blessé. D'autres travaux, notamment ceux de (Gratch et Marsella, 2004) se sont intéressés à la modélisation du comportement émotionnel d'un médecin face au cas médical classique de l'accompagnement des mourants par l'administration de médicaments. Dans le scénario présenté, le médecin souhaite prolonger le plus possible la vie de son patient âgé de 11 ans. La famille du patient quant à elle pourrait s'opposer à cette décision à cause de la souffrance que le patient va devoir endurer. Que fera le médecin dans ce cas-là ? Céder aux exigences de la famille ou ignorer leur décision et accomplir son devoir ?

Dans la littérature, nous trouvons aussi un ensemble de travaux sur le célèbre dilemme du tramway. Dans sa version originale, énoncé par (Foot, 1967), un tramway dont les freins sont défaillants roule sur la voie A et se dirige droit sur cinq ouvriers. La seule issue possible est d'actionner un levier pour dévier le tramway de sa trajectoire initiale pour le rediriger vers la voie B où se trouve un seul ouvrier. Que faire dans ce cas ? Sacrifier la personne sur la voie B pour sauver les cinq ou ne pas intervenir et laisser le tramway suivre son chemin ? Ce dilemme et ses variantes ont fait l'objet de plusieurs études sur papier (Hauser *et al.*, 2007), (Valdesolo et Desteno, 2006) et en environnements virtuels (Navarrete *et al.*, 2012), (Skulmowski *et al.*, 2014).

Dans les travaux cités jusqu'ici, l'approche adoptée pour la construction des dilemmes est dite « scriptée », c'est-à-dire, que l'écriture des situations par les auteurs se fait à l'avance, en amont de l'exécution de la simulation. A l'opposé, nous trouvons les approches génératives, qui consistent à générer dynamiquement des dilemmes au fur et à mesure qu'on avance dans la simulation. A notre connaissance, GADIN (Barber et Kudenko, 2007) est le seul système de l'état de l'art qui adopte une telle approche. GADIN est un moteur narratif interactif, qui confronte l'utilisateur, au fur et à mesure qu'il avance dans l'histoire, à des situations de dilemmes en utilisant des techniques de planification. L'histoire évolue en fonction des décisions de l'utilisateur face à ces dilemmes. Afin de les générer, les auteurs proposent cinq catégories identifiées selon les conséquences qu'aura l'action de l'utilisateur sur lui-même, ses amis et/ou ses ennemis : « Trahison », « Sacrifice », « Bien commun », « Mal commun » et « Faveur ».

### 2.1.2. Positionnement

L'approche scriptée permet de décrire en amont, d'une manière fine et précise, les dilemmes à présenter à l'utilisateur. L'utilisation de cette approche est pertinente lorsqu'il s'agit d'expérimentations en sciences humaines ou d'entrainement à des scénarios bien spécifiques qui ne nécessitent pas une variété de situations. Cependant elle reste non adaptée à nos besoins d'entrainement en situations critiques. En effet, nous souhaitons confronter l'apprenant à un ensemble de dilemmes qui varient en fonction des consignes pédagogiques. La nécessité d'une variabilité de situations ainsi que le besoin d'une liberté d'action de l'utilisateur dans l'environnement virtuel rendent difficiles voire impossible l'écriture exhaustive de tous les dilemmes possibles. Pour remédier à ce problème, GADIN propose une approche générative. Cependant, ce système souffre de deux limites majeures. Premièrement, la liberté d'action de l'utilisateur est réduite. En effet, face aux situations, une boite de dialogue s'offre à l'utilisateur où il peut choisir de faire ou ne pas faire une action. Deuxièmement, toutes les catégories proposées dépendent nécessairement des relations sociales (amis ou ennemis) qu'entretient l'utilisateur avec les autres personnages. Il est donc impossible de générer un dilemme si l'utilisateur est seul dans l'environnement ou lorsqu'il n'entretient pas de relations particulières avec les autres personnages.

A l'instar de GADIN, l'approche que nous proposons permet de générer dynamiquement des dilemmes sans avoir à les écrire en amont. Elle permet d'ajuster la criticité au fur et à mesure de la simulation, et donc d'adapter la nature des dilemmes à présenter, en fonction des actions de l'utilisateur et de son profil. Par ailleurs, elle repose sur une modélisation plus large des dilemmes qui ne se restreint pas aux relations sociales qu'entretient l'utilisateur avec les autres personnages.

## 3. Modélisation d'un dilemme

### 3.1. Définitions

Un dilemme est défini comme une situation où il n'existe pas de bonne solution. C'est une situation où l'individu est confronté à un choix difficile, sacrifiant, mettant en conflit dans certains cas, ses valeurs. Ce genre de situations peut être identifié par un humain. Par conséquent, un formateur, disposant d'une interface appropriée, pourrait facilement identifier une situation dilemmatique et la proposer à l'apprenant durant sa session d'entrainement. Cependant dans le cadre d'une approche de scénarisation dynamique, ce processus d'identification et de sélection de situations doit se faire automatiquement par un système informatique. Ce type d'approche soulève les questions suivantes :

- Comment modéliser informatiquement un dilemme ?
- Comment construire une sémantique associée ?
- Comment permettre à un moteur de scénarisation de générer dynamiquement un dilemme à partir de cette sémantique ?

La définition du Larousse, évoquée plus haut, met l'accent sur une propriété importante à laquelle répondent les situations dilemmatiques : les conséquences sont toujours négatives quel que soit le choix. Certains auteurs (Vallentyne, 1989) font une distinction entre les *« Obligation Dilemmas »* (que nous appellerons dilemmes d'obligation) et les *« Prohibition Dilemmas »* (que nous appellerons dilemmes de prohibition). Le premier type fait référence aux situations où toutes les actions sont obligatoires mais ne peuvent pas être toutes réalisées. Le deuxième type se réfère aux situations où toutes les actions sont prohibées mais au moins l'une d'entre elle doit être réalisée.

## 3.2. Conditions nécessaires

Dans les dilemmes d'obligation, la condition que doit satisfaire la situation est la suivante : les actions ne peuvent pas toutes être réalisées. Autrement dit, le choix présenté à l'agent doit être exclusif. En effet, si l'agent a la possibilité de réaliser toutes les actions, sans retombées négatives, le dilemme serait compromis. L'enjeu d'un système de scénarisation est de pouvoir prescrire un état du monde qui garantit cette condition. Pour ce faire, nous proposons de raisonner sur la contradiction entre les actions. Deux actions sont considérées contradictoires si elles sont incompatibles quand il faut les réaliser simultanément (p. ex. « avancer » et « reculer », « augmenter » et « diminuer »). Ceci permet de garantir que l'agent, lors de son choix face à une alternative, ne puisse pas choisir les deux issues.

Dans les dilemmes de prohibition, la condition nécessaire est que l'agent est obligé de réaliser au moins une action. L'enjeu pour un système de scénarisation est de pousser l'agent à faire un choix alors que chacune des actions a des conséquences négatives. Lorsque les actions sont prohibées, la seule issue positive pour l'agent serait de ne rien faire. Afin de garantir le dilemme, il faut écarter cette possibilité positive. Il faut donc que le non-choix conduise également à des conséquences négatives.

## 3.3. Formalisation

### 3.3.1. Dilemme d'obligation

Soit $a_1$ et $a_2$ deux actions qu'un agent AG peut réaliser. Pour certaines raisons, AG doit réaliser $a_1$. En d'autres termes, ne pas faire $a_1$ conduit à des conséquences négatives NC (présentées dans 3.4). Pour d'autres raisons, AG doit réaliser $a_2$. En d'autres termes, ne pas faire $a_2$ conduit à des conséquences négatives. La condition nécessaire dans un dilemme d'obligation est que l'agent peut réaliser soit l'action $a_1$ soit l'action $a_2$, mais pas les deux. Par conséquent, S est une situation de dilemme d'obligation si :

$$\neg a_1 \xrightarrow{Leads\ to} NC_{\neg a_1}$$
$$\neg a_2 \xrightarrow{Leads\ to} NC_{\neg a_2}$$
$$(a_1 \lor a_2) \land \neg(a_1 \land a_2)$$

### 3.3.2. Dilemme de prohibition

Soit $a_1$ et $a_2$ deux actions qu'un agent AG peut réaliser. Pour certaines raisons, AG ne doit pas réaliser $a_1$. En d'autres termes, faire $a_1$ conduit à des conséquences négatives NC (présentées dans 3.4). Pour d'autres raisons, AG ne doit pas réaliser $a_2$. En d'autres termes, faire $a_2$ conduit à des conséquences négatives. La condition nécessaire dans un dilemme de prohibition est que, malgré cela, l'agent doit réaliser au moins une action. Par conséquent, le non-choix (ne pas faire $a_1$ et ne pas faire $a_2$) conduit également à des conséquences négatives. S est alors une situation de dilemme de prohibition si :

$$a_1 \xrightarrow{Leads\ to} NC_{a_1}$$
$$a_2 \xrightarrow{Leads\ to} NC_{a_2}$$
$$\neg a_1 \wedge \neg a_2 \xrightarrow{Leads\ to} NC_{\neg a_1 \wedge \neg a_2}$$

## 3.4. Conséquences négatives

Dans le cadre de nos travaux, les conséquences négatives ne se résument pas seulement aux dommages causés suite à la réalisation d'une action. Elles incluent également d'autres types de conséquences. Nous distinguons en effet des conséquences négatives en termes de :

– Gravité : nombre de victimes, degré des blessures et/ou dégâts matériels (p. ex. percuter un piéton, un arbre ou un autre véhicule) ;

– Violations : infractions de code, atteinte aux valeurs morales, non-respect des normes, règles et instructions (p. ex. griller un feu rouge, désobéir à un supérieur) ;

– Perte de points : perte de points de performance, diminution du score (p. ex. perte de points de permis).

## 4. Architecture

Le système de scénarisation que nous proposons est intégré au sein de la plateforme logiciel HUMANS (Lourdeaux *et al.*, 2017). Il est composé d'un moteur de génération et d'un moteur de planification. Le premier est responsable de la génération des dilemmes. Il reçoit en entrée une consigne pédagogique de la part du module de diagnostic de l'apprenant. Cette consigne est composée (1) d'une intention pédagogique, comme vérifier, renforcer ou déstabiliser une compétence et (2) d'un degré de criticité. Elle est transformée ensuite en objectifs scénaristiques qui sont transmis au moteur de planification. Ce dernier est chargé d'identifier les ajustements scénaristiques adéquats, afin de diriger la simulation vers un état du monde qui répond aux objectifs scénaristiques donnés en entrée. Ces deux précédents modules communiquent avec un gestionnaire du monde qui est responsable de la mise à jour de l'état du monde. Il est également chargé de notifier l'environnement virtuel de tout changement dans le monde, afin de déclencher les comportements visuels correspondants. L'ensemble des modules du système de scénarisation manipule des modèles de connaissances que nous détaillons dans la prochaine section. Le schéma présenté en Figure 1 illustre cette architecture.

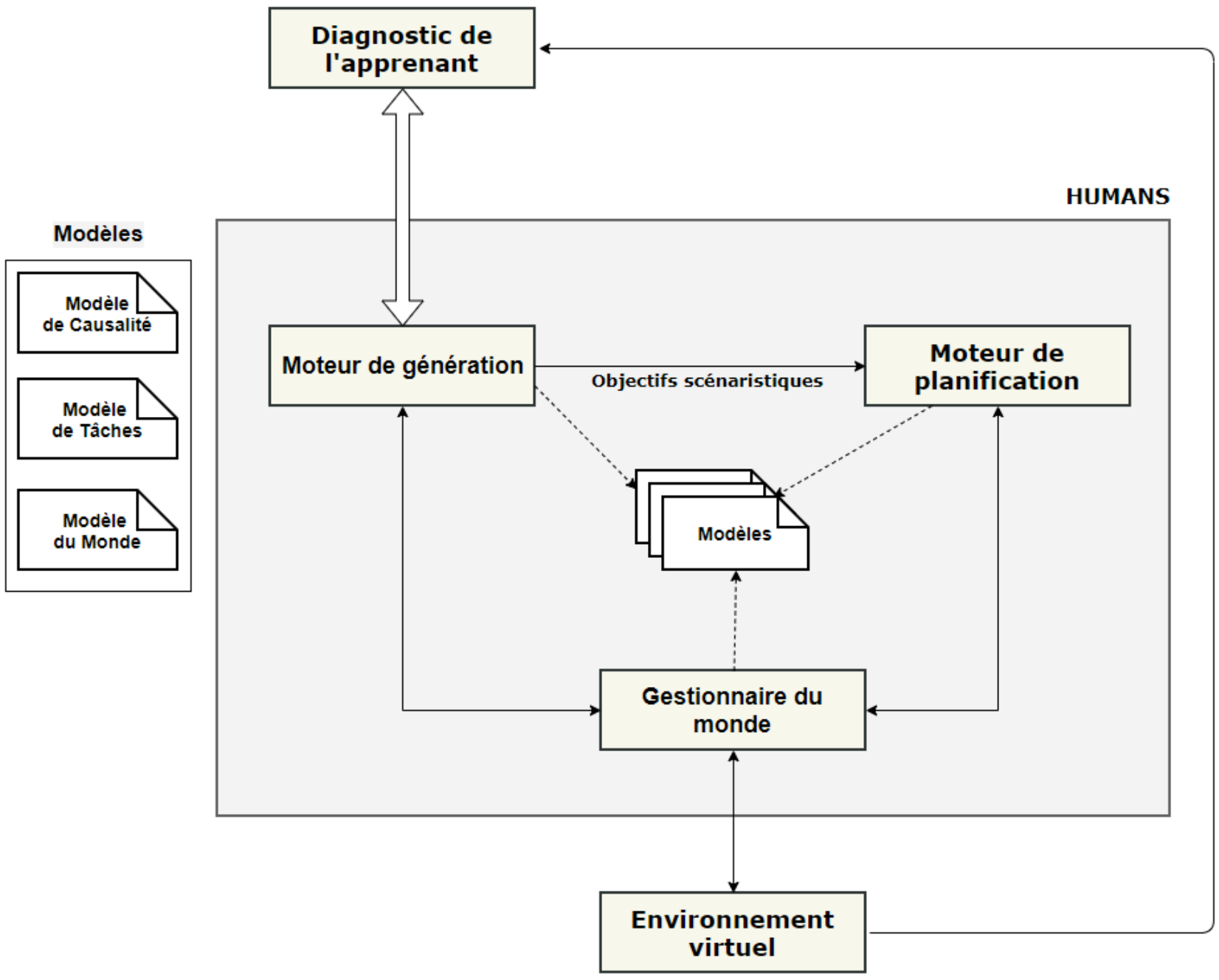

**Figure 1 · Architecture globale**

## 5. Modèles de connaissances

La conception d'environnements virtuels pour l'entraînement n'est pas du ressort des informaticiens seulement. C'est une tâche collective qui mobilise différentes personnes à savoir des formateurs, des ergonomes, des experts, etc. qui ne sont pas nécessairement informaticiens. Par conséquent, les connaissances spécifiques au domaine (p. ex. les objets, leurs propriétés, les actions) doivent être séparées de la représentation dans l'environnement virtuel 3D. Cette modularité facilite la conception collective de la simulation et permet aux experts de manipuler des outils et des modèles adaptés à leur domaine d'expertise. Nous pensons qu'au minimum trois modèles sont nécessaires, qu'ils soient considérés comme des vues d'un modèle central ou qu'ils soient des modèles distincts indépendants, l'essentiel c'est qu'ils doivent permettre de renseigner trois catégories majeures de connaissances : des connaissances sur le monde, des connaissances sur les tâches mais aussi des connaissances sur la causalité des événements – principalement lorsqu'il s'agit d'entraînement aux situations à risques. Ces connaissances peuvent être complétées par des connaissances pédagogiques et/ou didactiques. Celles-ci sont principalement utilisées par le module du diagnostic de l'apprenant qui est géré par nos partenaires dans le projet. Ce module ainsi que ces connaissances ne sont pas l'objet de cet article.

### 5.1. Modèle du monde

Le modèle du monde est destiné à être rempli par les experts du domaine qui ne sont pas nécessairement des informaticiens. Il doit donc être intelligible tout en étant interprétable par un système informatique. Ce modèle doit en plus offrir une large expressivité. Il doit permettre de renseigner les objets du monde, leurs propriétés ainsi que les relations entre eux, et de raisonner à différents niveaux d'abstraction, afin de permettre une plus grande variabilité de situations. Pour répondre à ces besoins, notre modèle s'appuie sur une représentation ontologique qui permet d'interroger la base de connaissances du monde afin d'extraire des informations pertinentes pour la scénarisation (p. ex. quels sont tous les feux de circulation qui se trouvent, au plus, à un rayon de 100m du véhicule de l'apprenant ?). Ces connaissances peuvent être couplées à des règles de fonctionnement pour permettre de gérer dynamiquement l'évolution du monde. Un exemple de modèle du monde est illustré dans la Figure 2.

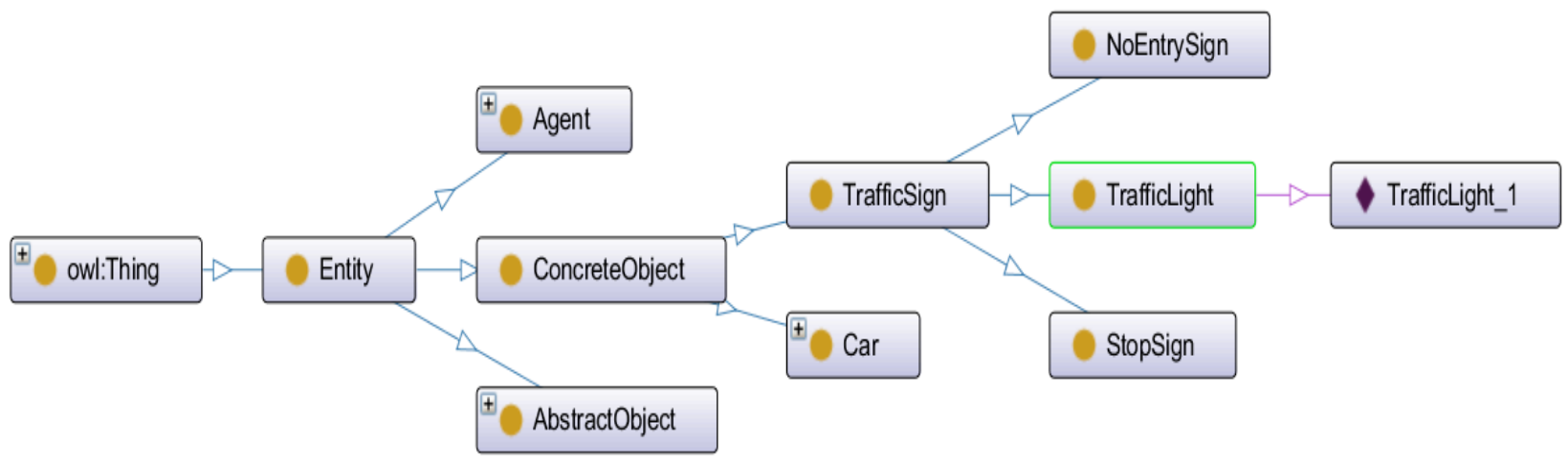

**Figure 2 · Fragment d'un modèle du monde**

## 5.2. Modèle de tâches

Le modèle de tâches est destiné à être rempli par des experts ergonomes. Par conséquent, il est utile qu'il repose sur une représentation qui opérationnalise des principes issus de langages utilisés en ergonomie cognitive (Sebillotte et Scapin, 1994), (van der Veer *et al.*, 1996). Par exemple, dans certaines représentations hiérarchiques, les tâches mères sont composées de sous-tâches filles sur plusieurs niveaux. Ces dernières sont liées par des constructeurs qui permettent de renseigner les relations logiques et temporelles entre elles. Les tâches peuvent avoir des préconditions et des postconditions exprimées, par exemple, par des agrégats d'assertions sur le monde sous la forme (*sujet prédicat objet*). Cette formulation nous parait très pertinente pour un couplage avec une représentation ontologique du monde. Certaines représentations distinguent, entre autres, les préconditions contextuelles et les préconditions favorables. Les premières sont les conditions qui rendent pertinente la réalisation de la tâche. Les dernières sont les conditions qui rendent la réalisation de la tâche préférable à d'autres. Les postconditions, quant à elles, appelées aussi conditions de satisfaction, renseignent sur l'état que le monde doit satisfaire pour qu'une tâche soit considérée comme réalisée. ACTIVITY-DL (Barot, 2014) est un langage qui regroupe les caractéristiques citées précédemment. C'est celui que nous utilisons dans le cadre de nos travaux. La Figure 3 montre une partie d'un modèle, représenté par ACTIVITY-DL, qui décrit une tâche « gérer feu rouge » composée de deux sous-tâches.

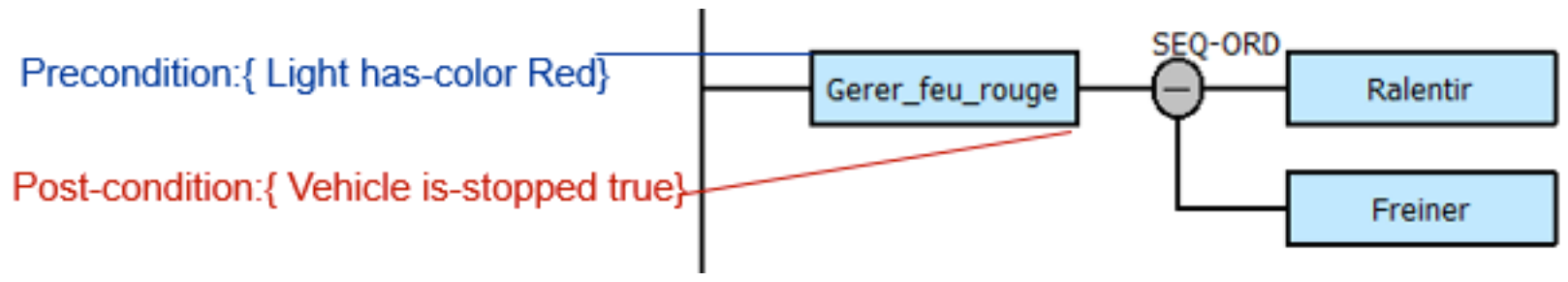

**Figure 3 · Fragment d'un modèle d'activité**

### 5.3. Modèle de causalité

Le modèle de causalité est inspiré du formalisme des nœuds papillons (Debray *et al.*, 2006) utilisé en analyse de risques. C'est un graphe acyclique et orienté qui exprime les chaines de causalité pertinentes de l'environnement. Les nœuds du graphe font référence aux événements. Ils sont reliés entre eux par des liens de subsomption ou de causalité. Ceci permet de déduire les chaines de causalité qui mènent aux conséquences négatives décrites dans la sous-section 3.4. Ils peuvent être étiquetés par tout type d'information pertinente. En analyse de risque par exemple, on s'intéresse au degré de gravité et/ou à la fréquence d'occurrence des événements (p. ex. gravité=2). Le modèle est aussi composé de connecteurs logiques « ET » et « OU » qui permettent aux auteurs de renseigner des événements causés par plus d'un événement. Par exemple, un « Accident » est causé par les événements « Véhicule roulant à grande vitesse » ET « Piéton qui traverse soudainement ». Aussi, un « Retrait de point permis » peut être causé par les événements « Griller un feu rouge » OU « Dépassement de vitesse autorisée ». Pour empêcher le déclenchement de certains événements, il existe des barrières de prévention. En effet, le déclenchement d'un événement est conditionné par la non-réalisation des barrières qui le précèdent, si elles existent. Ces barrières sont tout simplement des actions ou tâches du modèle de tâches. Pour bien comprendre ce modèle et ses éléments, nous proposons l'exemple présenté dans la Figure 4. Dans cet exemple, une « Violation du code de la route » est causé soit par « Griller un feu rouge », soit par « Griller un stop ». Ces deux événements risquent de se produire lorsque, l'apprenant s'approche, respectivement, du feu rouge ou du stop. Pour empêcher leur déclenchement, il peut activer les barrières correspondantes : « S'arrêter » ou « Marquer un temps d'arrêt ».

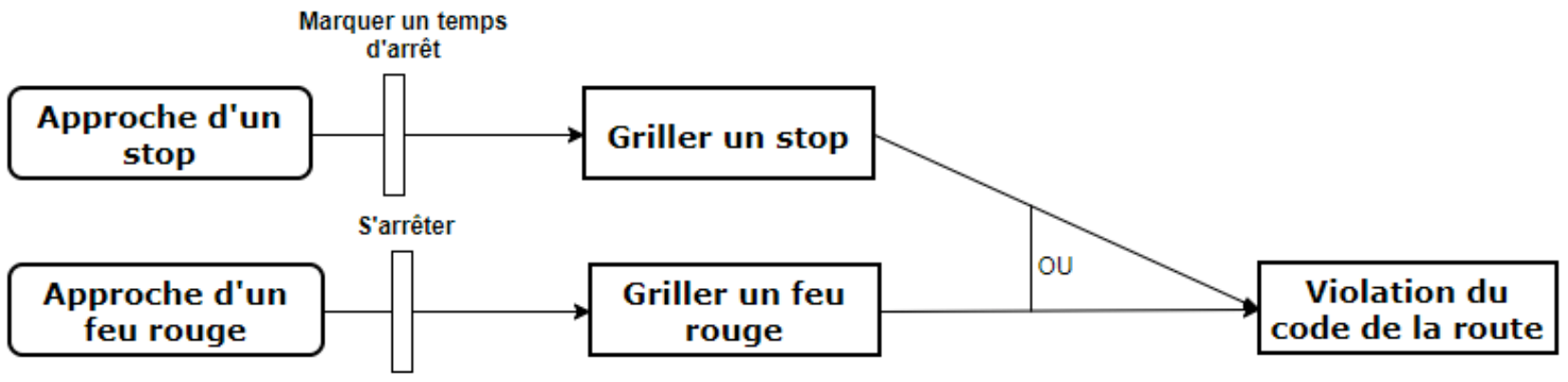

**Figure 4 · Fragment d'un modèle de causalité**

## 6. Génération de dilemmes

Nous avons décrit plus haut les différents types de dilemmes ainsi que les conditions que ces types doivent satisfaire. Par ailleurs, nous avons présenté les différents modèles de connaissances utilisés par le système de scénarisation. Dans cette section, nous détaillons les différentes étapes de génération illustrées dans la Figure 5. Nous expliquons comment le moteur de génération utilise ces domaines de connaissances, afin d'extraire les propriétés nécessaires et de permettre par conséquent l'émergence de dilemmes.

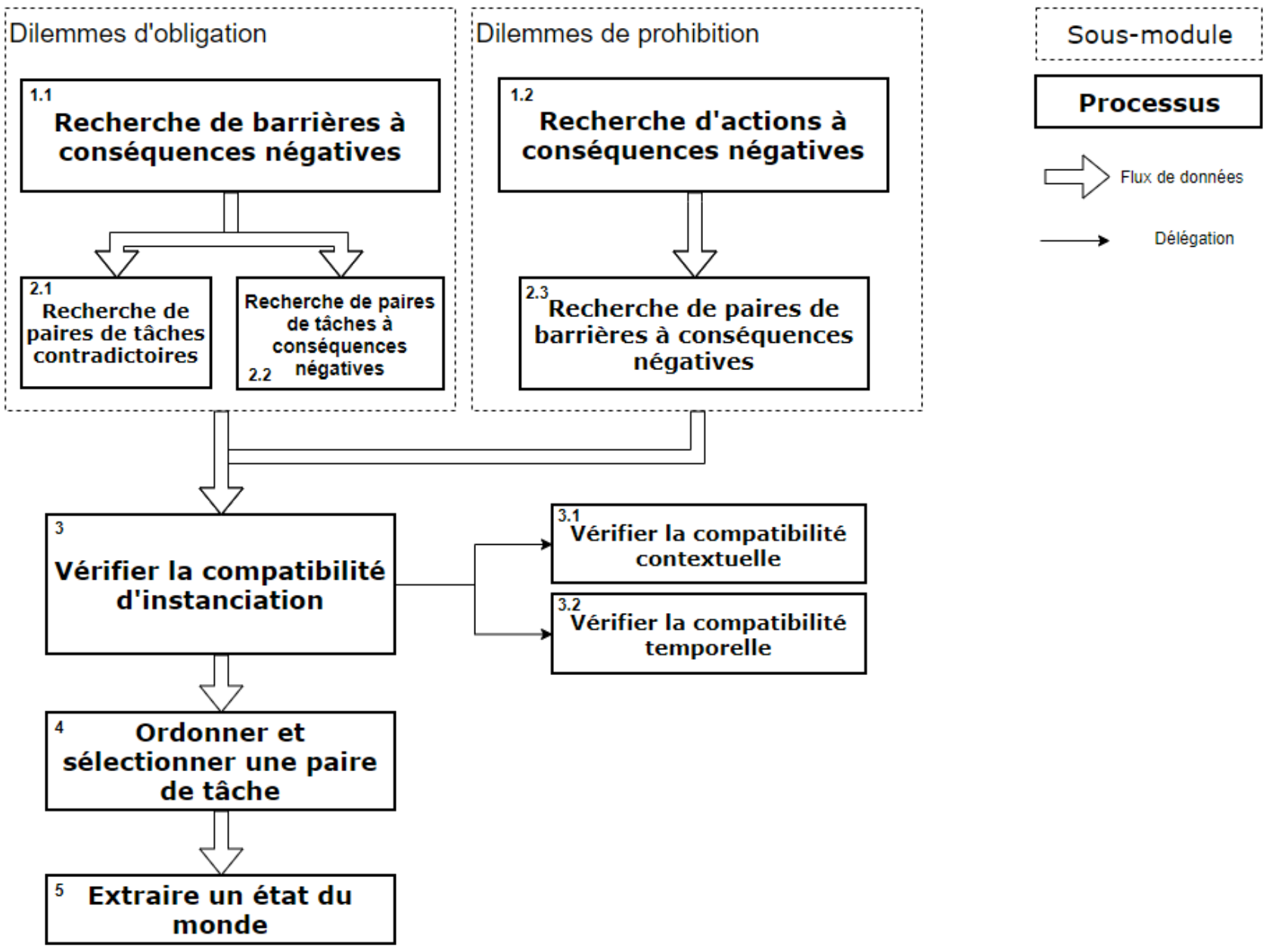

**Figure 5 · Etapes de génération de dilemmes**

### 6.1. Recherche d'actions/barrières à conséquences négatives

Pour la génération des dilemmes d'obligation, nous nous intéressons aux actions dont la non-réalisation conduit à des conséquences négatives. Cela correspond exactement à la définition d'une barrière dans le modèle de causalité. En effet, si une barrière n'est pas réalisée par l'agent, elle déclenche ses événements postérieurs. Par conséquent, rechercher les actions dont la non-réalisation conduit à des conséquences négatives revient à rechercher les barrières dont les événements postérieurs conduisent aux nœuds « Gravité », « Violations » et/ou « Points » (1.1). Pour garantir que ces conséquences puissent se déclencher, il ne doit pas y avoir d'autres barrières entre celle choisie et les nœuds conséquence, sinon l'agent pourrait éviter les conséquences négatives en activant l'une d'entre elles. Dans l'exemple présenté dans la Figure 6, la barrière $b_1$ est écartée car ses événements postérieurs ne mènent pas à des conséquences négatives. La barrière $b_3$ est également écartée car il existe une barrière entre elle et le nœud « Violations ». Au final, seules les barrières $b_2$ et $b_4$ sont retenues, car leurs événements postérieurs conduisent à des nœuds de conséquences négatives sans qu'il y ait aucune autre barrière dans le chemin qui mène à ces nœuds.

Pour générer les dilemmes de prohibition, nous nous intéressons à l'identification des actions dont la réalisation conduit à des conséquences négatives (1.2). Pour cela, nous recherchons toutes les actions qui sont liées aux nœuds « Gravité », « Violations » et/ou « Points ». Pareillement, il ne doit pas y avoir de barrière entre le nœud action et le nœud conséquence pour la raison évoquée plus haut. Dans l'exemple présenté dans la Figure 6, les actions $a_1$ et $a_2$ conduisent à des conséquences négatives. Toutefois, $a_1$ est écartée car il existe une barrière entre elle et le nœud conséquence.

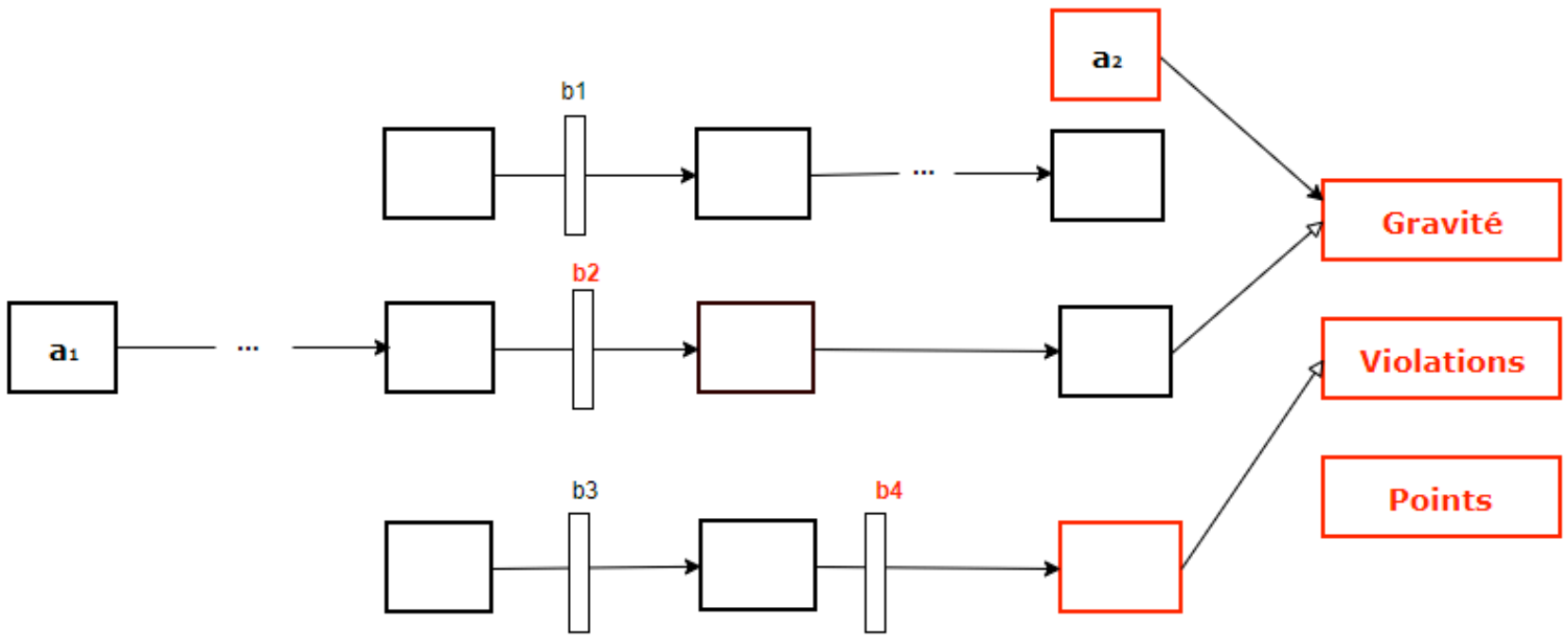

**Figure 6 · Sélection d'actions et de barrières**

## 6.2. Recherche de paires de tâches contradictoires (dilemme d'obligation)

L'extraction des paires de tâches potentiellement contradictoires (2.1) se fait grâce au modèle de tâches selon l'algorithme 1 (Figure 7). Le moteur de génération scanne, par paires, l'ensemble des barrières sélectionnées (tâches qui conduisent à des conséquences négatives si elles ne sont pas réalisées). Pour chaque paire, il examine les postconditions (conditions de satisfaction). Si elles sont incompatibles, alors la paire est nomologiquement incompatible (*i.e.* opposée par nature, p. ex. augmenter/diminuer). Nous considérons que deux conditions sont incompatibles si, pour le même couple (sujet prédicat), l'objet est différent (p. ex. « *Vehicle is-stopped true* » et « *Vehicle is-stopped false* »). Cet algorithme est généralisable à n tâches. Le système utilise dans ce cas une liste de tâches contradictoires au lieu de paires.

```
Algorithme 1: Recherche d'actions contradictoires
Function fetchContradictoryActions(Tasks ⊂ Barriers)
Begin
 PCT ← {} /* Ensemble des tâches contradictoires*/
  For all t1 ∈ Tasks do
   For all t2  ∈ Tasks do
    If ( t1.postCondition.subject = t2. postCondition.subject and
         t1.postCondition.predicate = t2. postCondition.predicate and
         t1.postCondition.object ≠ t2. postCondition.object )
     If {t1,t2} ∉ PCT then
         PCT ← PCT ∪ {{t1,t2}}
  Return PCT;
```

**Figure 7 · Algorithme de recherche d'actions contradictoires**

### 6.3. Recherche de paires de barrières à conséquences négatives (dilemme de prohibition)

La condition nécessaire dans un dilemme de prohibition est que l'agent doit choisir au moins une action. Nous avons expliqué que pour pousser l'apprenant à choisir il faudrait que le non-choix soit pénalisant aussi. En d'autres termes, ne réaliser aucune action conduirait aussi à des conséquences négatives. Pour identifier ce genre d'actions, le moteur de génération utilise le modèle de causalité. Il recherche les barrières qui sont liées par des portes « AND » (2.3) suivant l'algorithme 2 (Figure 8).

**Algorithme 2 :** *Recherche de barrières qui mènent à des conséquences négatives*

```
Function fetchBarriersConnectedWithAND(Barriers)
Begin
 PB ← {} /* Ensemble des paires de barrières */
  For all b1 ∈ Barriers do
   For all b2 ∈ Barriers do
    If ( {"AND"} ∈ CommonDescendant(b1,b2) )
      PB ← PB ∪ {{b1,b2}}
  Return PB;
End
```

**Figure 8 · Algorithme de recherche de barrières qui mènent à des conséquences négatives**

### 6.4. Vérification de la compatibilité d'instanciation

Afin de garantir que le système de scénarisation puisse instancier une situation mettant en jeu une paire de tâches, ces dernières doivent être contextuellement compatibles (3.1). C'est-à-dire qu'elles doivent disposer de préconditions compatibles. Considérons par exemple les tâches « Ouvrir Porte » et « Fermer Porte » dont les préconditions sont respectivement *« Door is-open false »* et *« Door is-open true »* et dont les postconditions sont respectivement *« Door is-open true »* et *« Door is-open false »*. Ces tâches sont nomologiquement incompatibles (analyse des postconditions), par conséquent elles pourraient éventuellement être utilisées pour la génération d'un dilemme ; en revanche, elles sont aussi contextuellement incompatibles (analyse des préconditions). La paire serait donc écartée. En effet, l'instanciation d'une telle situation nécessiterait que la porte soit ouverte et fermée en même temps, ce qui est impossible.

En plus de la compatibilité contextuelle, les paires de tâches doivent satisfaire la compatibilité temporelle (3.2). C'est-à-dire que la réalisation d'une tâche doit être indépendante de l'autre. Dans le modèle de tâches, les tâches sont connectées par les constructeurs temporels de leur tâche parente. Le fait que deux tâches soient liées par un constructeur « séquentiel » implique que la réalisation d'une tâche est une précondition de l'autre. Par conséquent, seules les tâches dont les ancêtres communs ont des constructeurs « parallèle » ou « indépendant » sont maintenues.

### 6.5. Sélection de paires de tâches

A cette étape, le moteur de génération dispose de deux listes de paires de tâches. L'une correspond aux dilemmes d'obligation, tandis que l'autre correspond aux dilemmes de prohibition. Afin de déterminer les paires de tâches les plus pertinentes (4), le moteur établit un ordre selon un score calculé à partir de plusieurs contraintes. Les paires de tâches les plus pertinentes sont celles qui répondent au mieux aux contraintes pédagogiques, établies par le module du diagnostic de l'apprenant, et aux contraintes scénaristiques, qui sont plutôt liées aux restrictions imposées par l'état actuel du monde. Les contraintes pédagogiques sont définies par :

- une gravité maximale à ne pas dépasser,
- une gravité minimale à respecter,
- une différence de gravité entre les tâches d'une paire (on peut penser que plus cette différence tend vers zéro, plus le dilemme est difficile),
- un type de conséquences si l'on souhaite générer un type spécifique de dilemme, à savoir les dilemmes moraux.

Les contraintes scénaristiques, quant à elles, sont définies par la probabilité d'instanciation des événements (liée à la disponibilité des entités nécessaires au déclenchement des événements), d'une part, et par la temporalité des événements, d'autre part. La paire de tâches idéale est celle qui répond parfaitement aux deux types de contraintes. En pratique, ce n'est pas toujours le cas. Par conséquent, parfois on peut vouloir qu'un dilemme réponde absolument aux contraintes pédagogiques même si les chances qu'il soit effectivement instanciable sont infimes. A l'inverse, on peut vouloir qu'un dilemme soit absolument instanciable même s'il ne répond pas exactement aux contraintes pédagogiques. Pour permettre de telles préférences, il est possible d'attribuer un poids à chaque type de contraintes par le formateur ou automatiquement par le module de diagnostic de l'apprenant. A partir de ces poids, et selon les tâches de chaque paire, un score est calculé et attribué pour chacune d'elle. Ce score permet d'ordonner les paires de tâches, et par conséquent de proposer les dilemmes les plus pertinents et les plus satisfaisants vis-à-vis des contraintes pédagogiques et scénaristiques.

## 7. Preuve de concept

L'approche décrite dans cet article a fait l'objet d'une première implémentation en environnement virtuel. Nous avons développé une première version de notre système de scénarisation ainsi qu'un environnement virtuel pour la conduite automobile (Figure 9) en utilisant le moteur de rendu Unity3D (https://unity3d.com/fr). L'environnement 3D est composé de bâtiments, de panneaux de signalisation, des feux de circulation ainsi que des agents autonomes (véhicules et piétons). Le joueur contrôle un véhicule en première personne en utilisant une configuration clavier-souris ou en utilisant un kit volant-pédale adapté.

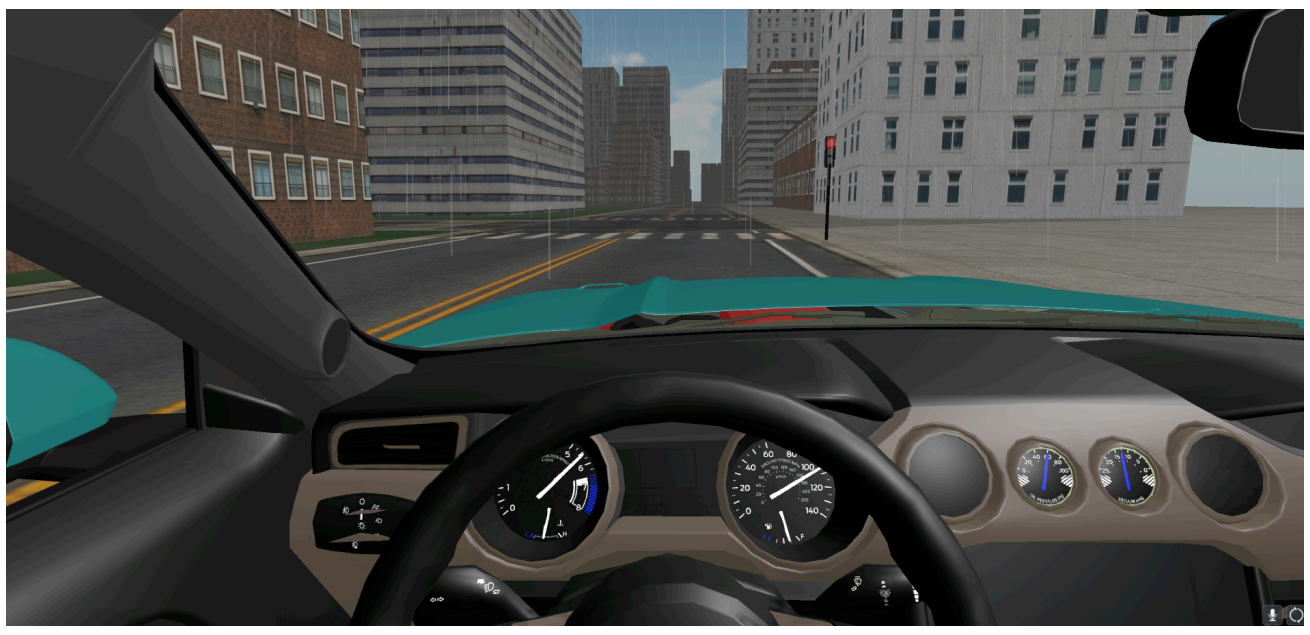

**Figure 10 · Capture d'écran de l'environnement**

Pour cette version du système nous avons utilisé le modèle de tâches illustré dans la Figure 10. Le modèle est volontairement simplifié pour des raisons de clarté. Il décrit trois tâches indépendantes : « *Handle_aquaplaning* », « *Handle_red_light* » et « *Handle_stop* ».

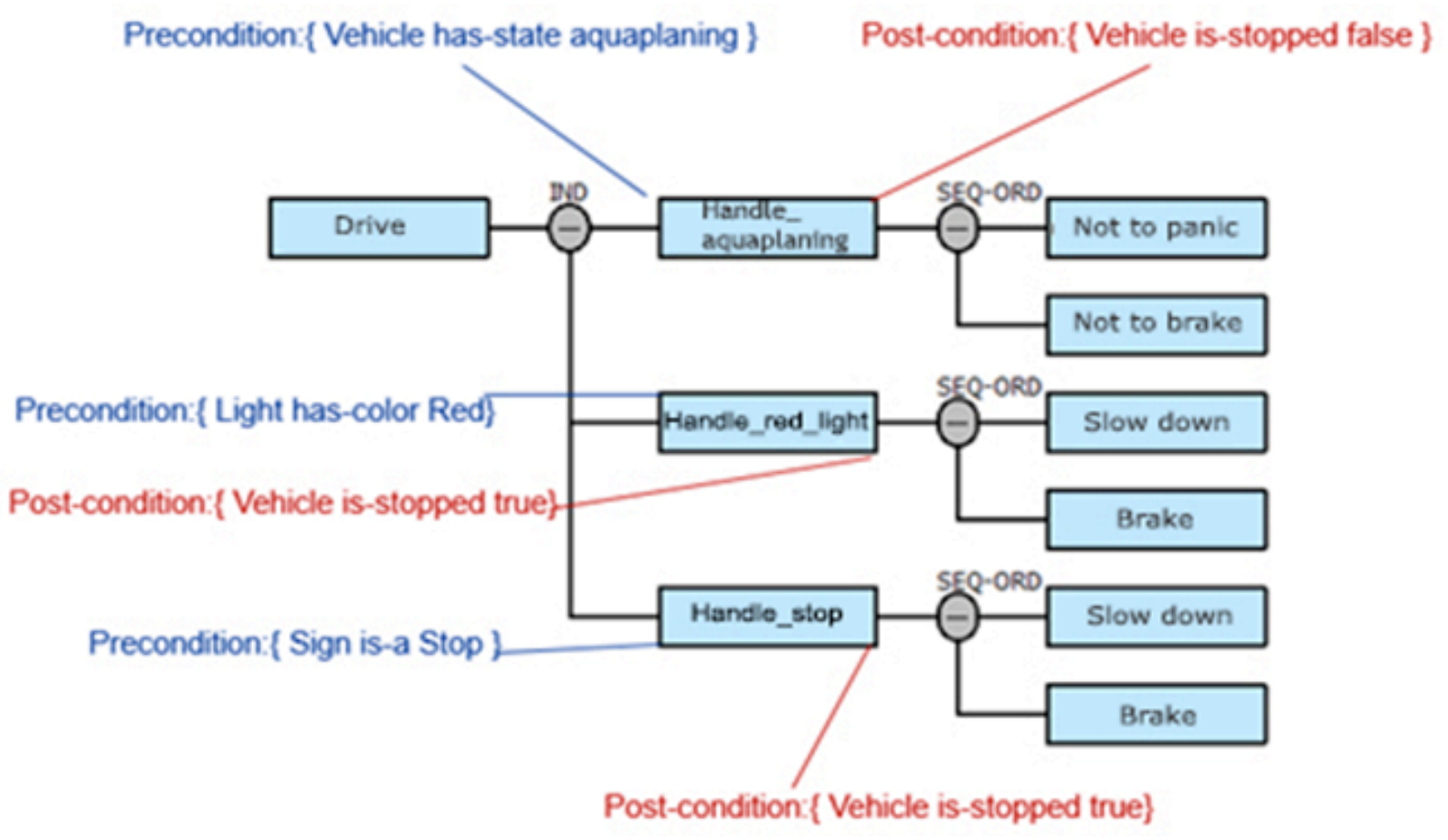

**Figure 11 · Modèle de tâches**

Nous avons aussi utilisé le modèle de causalité illustré dans la Figure 11. Il décrit un ensemble d'événements comme par exemple « *Running a Stop Sign* » et « *Running a Red Light* » qui conduisent à un « *Highway Code Violation* ». Ce dernier peut aussi être causé par les nœuds « *Driving* » et « *Answer a phone call* ».

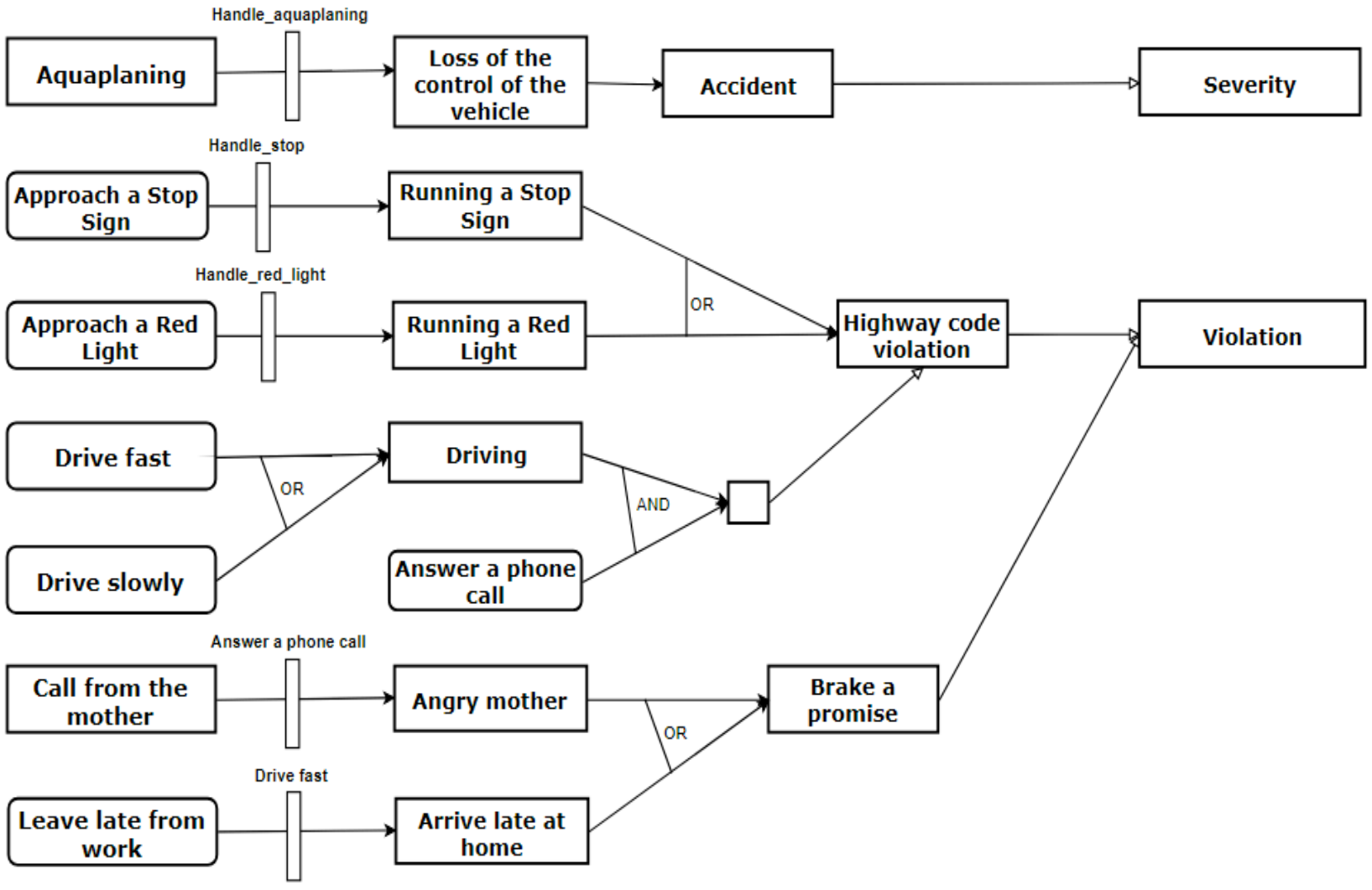

**Figure 12 · Modèle de causalité**

Les différentes étapes de l'exécution de l'algorithme de génération de dilemme (Figure 5) sont détaillées ci-dessous.

**Etape 1.1 : détermination des barrières avec des conséquences négatives (dilemmes d'obligation)**

D'après le modèle de causalité, les barrières qui, si elles ne sont pas activées, pourraient mener à des conséquences négatives sont *"Handle_stop", "Handle_red_light", "Handle_aquaplaning", "Answer a phone call"* et *"Drive fast"*.

**Etape 1.2 : détermination des actions avec des conséquences négatives (dilemmes de prohibition)**

D'après le modèle de causalité, les actions qui pourraient mener à des conséquences négatives sont *"Approach a Stop sign", "Approach a Red light", "Drive fast", "Drive slowly", "Answer a phone call"* et *"leave late from work"*. Les deux premières sont écartées car il existe des barrières entre elles et le nœud conséquence.

**Etape 2.1 : détermination des paires de tâches contradictoires**

En parcourant la liste retournée à la fin de l'étape 1.1, le moteur de génération recherche les tâches contradictoires en s'appuyant sur le modèle de tâches. Il retourne les paires suivantes :

- Paire 1 : {*"Handle_stop", "Handle_aquaplaning"*},
- Paire 2 : {*"Handle_red_light", "Handle_aquaplaning"*}.

En effet, les postconditions des tâches, pour les deux paires, sont incompatibles (*« Vehicle is-stopped false »* vs *« Vehicle is-stopped true »*). La paire {*« Handle_stop », « Handle_red_light »*} est rejetée car les postconditions sont compatibles (*« Vehicle is-stopped true »* vs « *Vehicle is-stopped true* »).

**Etape 2.3 : détermination des paires de barrières avec conséquences négatives**

A partir de la liste retournée à la fin de l'étape 1.2, le moteur de génération cherche, par paire, les barrières qui conduisent à des conséquences négatives si elles sont activées toutes les deux. Dans cet exemple, le moteur ne retourne aucune paire car il n'existe pas de barrières qui ont un « AND » comme nœud descendant commun.

**Etape 3 : vérification de la compatibilité d'instanciation**

L'ancêtre commun des tâches de la Paire 1 est la tâche « *Drive* » dont le constructeur est le constructeur temporel indépendant (IND). C'est le cas aussi des tâches de la Paire 2. Par ailleurs, les préconditions des tâches de Paire 1 sont compatibles (*« Sign is-a Stop »* vs *« Vehicle has-state aquaplaning »*). De la même façon, pour Paire 2, les préconditions des tâches sont compatibles. Par conséquent, Paire 1 et Paire 2 sont toutes les deux retenues car leurs tâches sont contextuellement et temporellement compatibles.

**Etape 4 : Classer et sélectionner**

A ce stade, deux paires sont candidates. Les dilemmes générés par ces paires seraient les suivants.

– Paire 1 : dans une situation d'aquaplanage, le conducteur doit-il freiner pour respecter le stop au risque de perdre le contrôle de son véhicule ou doit-il éviter de freiner, en grillant ainsi le stop, afin de garder le contrôle du véhicule ?

– Paire 2 : dans une situation d'aquaplanage, le conducteur doit-il freiner pour respecter le feu au risque de perdre le contrôle de son véhicule ou doit-il éviter de freiner, en grillant ainsi le feu, afin de garder le contrôle du véhicule ?

Dans cette version de la preuve de concept, le calcul des scores des paires de tâches n'a pas été implémenté. La sélection de la meilleure paire s'est faite de manière aléatoire. Dans une version plus aboutie qui prendrait en compte les contraintes pédagogiques et scénaristiques, si on suppose qu'il y a dix fois plus de feux rouges que de panneaux stop, le moteur de génération pourrait préférer Paire 2 à Paire 1 car il y aurait plus de chance qu'une situation mettant en jeu les tâches de Paire 2 se produise. Pour la suite de cet exemple, nous supposerons alors que cette dernière a été choisie.

**Etape 5 : Extraction d'un état du monde**

Finalement le moteur de génération extrait un état du monde à partir de la paire de tâches sélectionnée. Cet état consiste en une agrégation des préconditions de ces tâches. Pour Paire 2, c'est {(*Vehicle has-state aquaplaning*) AND (*Light has-color Red*)}. Ces préconditions, qui correspondent à un état but sont transmises au moteur de planification qui se charge de diriger la simulation vers une situation où se trouve un feu rouge et où le véhicule du conducteur est en aquaplanage. Ensuite, l'environnement virtuel se charge d'instancier visuellement la situation.

## 8. Evaluation et résultats

Afin d'évaluer notre moteur de génération de dilemmes, nous avons mené une première évaluation sans environnement virtuel. Cette évaluation consistait en un questionnaire en ligne où les participants devaient répondre à un ensemble de questions par rapport à différentes situations de conduite. Le questionnaire portait sur neuf situations de conduite automobile décrites par deux phrases au maximum. Parmi ces situations, six d'entre elles correspondaient à des situations générées par notre système. Elles impliquaient les paires suivantes :

- {« Respecter Stop », « Gérer Aquaplanage »},
- {« Respecter Feu rouge », « Gérer Aquaplanage »},
- {« Gérer Voiture qui colle », « Respecter Stop »},
- {« Gérer Voiture qui colle », « Gérer Traversée Piéton »},
- {« Gérer Voiture qui colle », « Respecter Feu Rouge »},
- {« Gérer Aquaplanage », « Gérer Traversée Piéton »}.

Les trois autres correspondaient à des situations dites normales :

- {« Respecter Feu Rouge », « Gérer Route Dégagée »},
- {« Respecter Sens Interdit », « Suivre Conseil du Passager »},
- {« Gérer Feu Vert », « Gérer Traversée d'un Piéton ».

Les neuf situations ont été présentées dans un ordre aléatoire. Pour chacune d'elles, les participants devaient répondre aux questions suivantes :

- Que feriez-vous dans cette situation ?
- Avez-vous hésité avant de prendre votre décision ? Pourquoi ?
- Pensez-vous qu'il existe une solution sans conséquences négatives ?
- Pensez-vous qu'il existe une bonne solution ?

Le but de cette évaluation était de savoir si les participants retrouvaient dans les situations générées les propriétés des situations dilemmatiques. Nous avons eu un total de 67 participants. La Figure 13 présente les résultats recueillis.

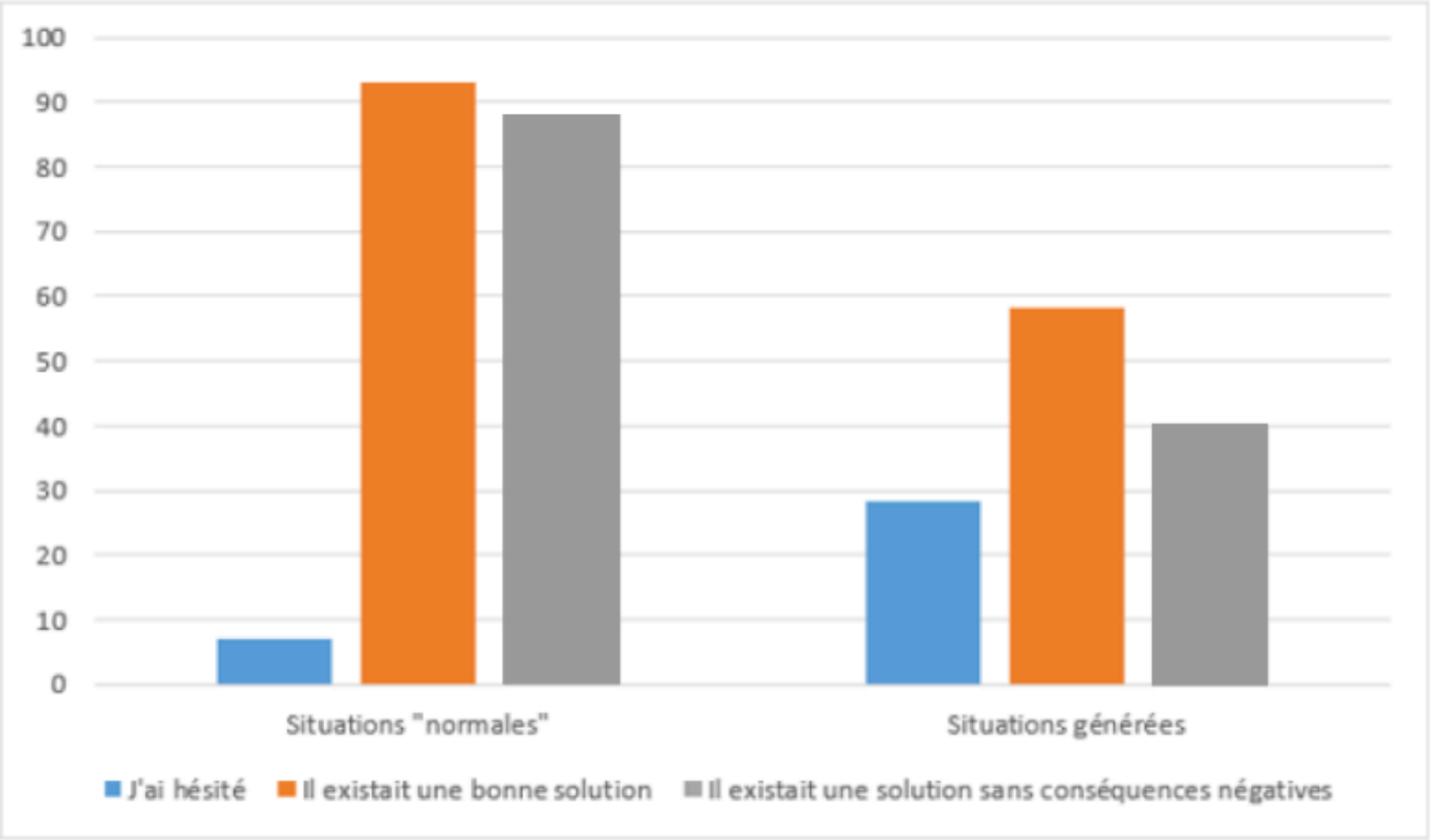

**Figure 13 · Résultats de l'évaluation**

Nous avons constaté que les participants étaient plus hésitants dans les situations générées que dans les situations normales. Mais nous nous attendions à une valeur d'hésitation plus élevée. En analysant les réponses des participants, il s'est avéré que ceci était dû à quelques éléments qui n'étaient pas pris en compte par notre système de génération, ce qui a permis à certaines décisions d'être plus ou moins simples et immédiates (p. ex. « *la plupart des véhicules de nos jours sont équipés de technologies qui font éviter les pertes de contrôle lors d'aquaplanage. Donc, devant un feu rouge je freine* »). Dans les situations générées, 60% des participants ont estimé qu'il n'y avait pas de solution sans conséquences négatives, tandis que pour les situations normales 12% seulement ont estimé cela. Par ailleurs, 93% des participants ont déclaré qu'il existait une bonne solution lorsqu'il s'agissait de situations normales contre 58% lorsqu'il s'agissait de situations générées. Par conséquent, nous pouvons en déduire que notre système a pu générer des situations qui, comparées à des situations normales, étaient plus compliquées en termes de prise de décision, étaient perçues comme ayant des conséquences négatives quel que soit le choix et ne présentaient pas de bonne solution, ce qui correspond aux situations dilemmatiques.

## 9. Conclusion

Dans cet article, nous avons proposé une modélisation informatique des dilemmes. Nous avons également proposé différents algorithmes qui permettent de raisonner sur des modèles de connaissances afin d'extraire les propriétés nécessaires pour l'émergence des situations dilemmatiques, tout en veillant à respecter au mieux les contraintes pédagogiques issues du diagnostic et les contraintes scénaristiques liées aux restrictions issues de l'état du monde. Ces travaux ont été validés par une preuve de concept en conduite automobile qui a mis en exergue la capacité du moteur de génération à générer des dilemmes sans qu'ils soient explicitement décrits à l'avance. Nous avons également conduit une évaluation préliminaire sous forme textuelle qui a permis d'observer que, comparé aux situations normales, les situations générées n'incluaient pas de bonne solution et que toutes les issues possibles étaient négatives. Ces résultats nous laissent optimistes pour une évaluation en environnement virtuel.

Face à un dilemme, chaque individu agira selon son propre système de valeurs (Williams, 1968). Lorsqu'aucune valeur ne prend le dessus sur l'autre, cela produit un conflit de valeurs. Pour être capable de générer de telles situations, la prise en compte du profil moral de l'apprenant est nécessaire. Une brève revue de la littérature nous laisse penser que la théorie des valeurs universelles de (Schwartz, 2006) serait un cadre théorique pertinent pour la modélisation du profil moral de l'apprenant. Cette théorie, construite grâce à des données issues de 67 pays, définit une typologie de valeurs structurées dans une représentation circulaire qui rend compte des relations de compatibilité et d'opposition des unes avec les autres. Pour nos travaux futurs, nous visons à opérationnaliser cette théorie et à intégrer par la suite le profil moral dans le processus de génération. L'objectif est de pouvoir produire des dilemmes moraux personnalisés selon le profil de chaque apprenant.